\documentclass[copyright,noncommercial]{eptcs}
\providecommand{\volume}{5}
\providecommand{\anno}{2009}
\providecommand{\firstpage}{13}
\providecommand{\eid}{2}
\providecommand{\event}{Y. Deville and C. Solnon (Ed.): Sixth International Workshop\\
 on Local Search Techniques in Constraint Satisfaction (LSCS'09).}
\usepackage{breakurl}        
\usepackage{color}
\usepackage{amsmath}
\usepackage{fancyvrb}
\usepackage{algorithmic}
\usepackage{algorithm}
\usepackage[british]{babel}
\usepackage{vaucanson-g}

\newtheorem{definition}{Definition}

\newcommand{\tuple}[1]{\langle #1 \rangle}

\newcommand{\Automaton}{\mathit{automaton}}
\newcommand{\Element}{\mathit{element}}
\newcommand{\DFA}{\mathit{DFA}}
\newcommand{\GCC}{\mathit{gcc}}
\newcommand{\MDD}{\mathit{mdd}}
\newcommand{\Pattern}{\mathit{pattern}}
\newcommand{\Regular}{\mathit{regular}}
\newcommand{\Stretch}{\mathit{stretch}}

\newcommand{\IsAssigned}{\leftarrow}
\newcommand{\Not}{\mathbf{not~}}
\newcommand{\Procedure}[1]{\STATE {\bf procedure}~#1}

\newcommand{\To}{{\bf ~to~}}

\newcommand{\True}{\mathbf{true}}
\newcommand{\False}{\mathbf{false}}

\newcommand{\calcSegment}{\mathit{calcSegment}}

\newcommand{\Vio}{\mathit{Violation}}

\title{Toward an \emph{automaton} Constraint for Local Search}

\author{
  Jun He \qquad\qquad Pierre Flener \qquad\qquad Justin Pearson
  \institute{
    Department of Information Technology \\
    Uppsala University,
    Box 337, SE -- 751 05 Uppsala, Sweden \\
  }
  \email{Firstname.Lastname@it.uu.se}
}

\def\titlerunning{Toward an $automaton$ Constraint for Local Search}
\def\authorrunning{J. He, P. Flener, and J. Pearson}

\begin{document}

\maketitle

\begin{abstract}
  We explore the idea of using finite automata to implement new
  constraints for local search (this is already a successful technique
  in constraint-based global search). We show how it is possible to
  maintain incrementally the violations of a constraint and its
  decision variables from an automaton that describes a ground checker
  for that constraint. We establish the practicality of our approach
  idea on real-life personnel rostering problems, and show that it is
  competitive with the approach of~\cite{Pralong:regular}.
\end{abstract}

\section{Introduction}

When a high-level constraint programming (CP) language lacks a
(possibly global) constraint that would allow the formulation of a
particular model of a combinatorial problem, then the modeller
traditionally has the choice of (1)~switching to another CP language
that has all the required constraints, (2)~formulating a different
model that does not require the lacking constraints, or
(3)~implementing the lacking constraint in the low-level
implementation language of the chosen CP language. This paper
addresses the core question of facilitating the third option, and as a
side effect often makes the first two options unnecessary.

The user-level extensibility of CP languages has been an important
goal for over a decade. In the traditional global search approach to
CP (namely heuristic-based tree search interleaved with propagation),
higher-level abstractions for describing new constraints include
indexicals \cite{ccfd}; (possibly enriched) deterministic finite
automata (DFAs) via the $\Automaton$ \cite{Beldiceanu:automata} and
$\Regular$ \cite{Pesant:seqs} generic constraints; and multi-valued
decision diagrams (MDDs) via the $\MDD$ \cite{Yap:mdd} generic
constraint. Usually, a generic but efficient propagation algorithm
achieves a suitable level of local consistency by processing the
higher-level description of the new constraint. In the more recent
local search approach to CP (called constraint-based local search,
CBLS, in \cite{Comet:book}), higher-level abstractions for describing
new constraints include invariants \cite{LocalizerCP97}; a subset of
first-order logic with arithmetic via combinators \cite{CometCP04} and
differentiable invariants \cite{CometInvariants}; and existential
monadic second-order logic for constraints on set decision variables
\cite{ASTRA:Constraints07:local}. Usually, a generic but incremental
algorithm maintains the constraint and variable violations by
processing the higher-level description of the new constraint.

In this paper, we revisit the description of new constraints via
automata, already successfully tried within the global search approach
to CP~\cite{Beldiceanu:automata,Pesant:seqs}, and show that it can
also be successfully used within the local search approach to CP. The
significance of this endeavour can be assessed by noting that
$108$ of the currently $313$ global constraints in the \emph{Global
  Constraint Catalogue} \cite{GC-catalogue} are described by DFAs that
are possibly enriched with counters and conditional
transitions~\cite{Beldiceanu:automata} (note that DFA generators can
easily be written for other constraints, such as the $\Pattern$
\cite{Pesant:pattern} and $\Stretch$~\cite{Pesant:stretch}
constraints, taking the necessarily ground parameters as inputs), so
that all these constraints will instantly become available in CBLS
once we show how to implement fully the enriched DFAs that are
necessary for some of the described global constraints.

The rest of this paper is organised as follows. In
Section~\ref{sect:viol}, we present our algorithm for incrementally
maintaining both the violation of a constraint described by an
automaton, and the violations of each decision variable of that
constraint. In Section~\ref{sect:exp}, we present experimental results
establishing the practicality of our results, also in comparison to
the prior approach of~\cite{Pralong:regular}. Finally, in
Section~\ref{sect:concl}, we summarise this work and discuss related
as well as future work.

\section{Incremental Violation Maintenance with Automata}
\label{sect:viol}

In CBLS, three things are required of an implemented constraint: a
method for calculating the violation of the constraint and each of its
decision variables for the initial assignment (initialisation); a
method for computing the differences of these violations upon a
candidate local move (differentiability) to a neighbouring assignment;
and a method for incrementally maintaining these violations when an
actual move is made (incrementality). Intuitively, the higher the
violation of a decision variable, the more can be gained by changing
the value of that decision variable. It is essential to maintain
incrementally the violations rather than recomputing them from scratch
upon each local move, since by its nature a local search procedure
will try many local moves to find one that ideally reduces the
violation of the constraint or one of its decision variables.

\begin{figure}[t]
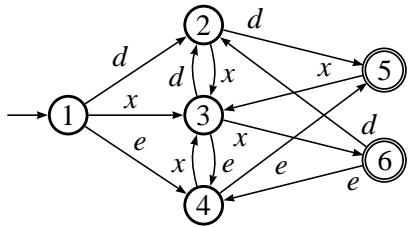

  \centering
  \VCDraw[1]{%
    \begin{VCPicture}{(0,-0.7)(8,4)}
      \State[1]{(0,2)}{1} 
      \State[2]{(3,4)}{2} \State[3]{(3,2)}{3} \State[4]{(3,0)}{4}
      \FinalState[5]{(7,3)}{5} \FinalState[6]{(7,1)}{6}
      \EdgeL{1}{2}{d} \EdgeL{1}{3}{x} \EdgeL{1}{4}{e}
      \ArcL[0.6]{2}{3}{x} \ArcL{3}{2}{d}
      \ArcL{4}{3}{x} \ArcL[0.6]{3}{4}{e}
      \EdgeL[0.2]{2}{5}{d} \EdgeR[0.15]{3}{6}{x} \EdgeR[0.35]{4}{5}{e}
      \EdgeR[0.25]{5}{3}{x}
      \EdgeR[0.06]{6}{2}{d} \EdgeL[0.1]{6}{4}{e}
      \Initial{1}
    \end{VCPicture}
  }
  \caption{An automaton for a simple work scheduling constraint}
  \label{fig:automaton}
\end{figure}

Our running example is the following, for a simple work scheduling
constraint. There are values for two work shifts, day ($d$) and
evening ($e$), as well as a value for enjoying a day off ($x$). Work
shifts are subject to the following three conditions: one must take at
least one day off before a change of work shift; one cannot work for
more than two days in a row; and one cannot have more than two days
off in a row. A DFA for checking ground instances of this constraint
is given in Figure~\ref{fig:automaton}. The start state~$1$ is marked
by a transition entering from nowhere, while the success states~$5$
and $6$ are marked by double circles. Missing transitions, say from
state~$2$ upon reading value~$e$, are assumed to go to an implicit
failure state, with a self-looping transition for every value (so that
no success state is reachable from it).

\subsection{Violations of a Constraint}

To define and compute the violations of a constraint described by an
automaton, we first introduce the notion of a segmentation of an
assignment:

\begin{definition}[Segmentation]
  Given an assignment $V=\tuple{d_1,\dots,d_n}$, a \emph{segmentation}
  is a possibly empty sequence of non-empty sub-strings (referred to
  here as \emph{segments}) $\sigma_1,\dots,\sigma_\ell$ of $d_1 \cdots
  d_n$ such that for each $\sigma_j = d_p \cdots d_q$ and
  $\sigma_{j+1} = d_r \cdots d_s$ we have that $r > q$.
\end{definition}

For example, a possible segmentation of the assignment
$V=\tuple{x,e,d,e,x,x}$ is $\tuple{x,e}, \tuple{e,x,x}$; note that the
third character of the assignment is not part of any segment. In
general, an assignment has multiple possible segmentations. We are
interested in segmentations that are accepted by an automaton, in the
following sense:

\begin{definition}[Acceptance]
  Given an automaton and an assignment $V=\tuple{d_1,\dots,d_n}$, a
  segmentation $\sigma_1,\dots,\sigma_{\ell}$ is \emph{accepted} by
  the automaton if there exist strings $\alpha_1,\dots,
  \alpha_{\ell+1}$, where only $\alpha_1$ and $\alpha_{\ell+1}$ may be
  empty, such that the concatenated string
  \[
    \alpha_1 \cdot \sigma_1 \cdot \alpha_2 \cdot
    ~\cdots~ \cdot \alpha_\ell \cdot \sigma_\ell \cdot \alpha_{\ell +1}
  \]
  is accepted by the automaton.
\end{definition}

For example, given the automaton in Figure~\ref{fig:automaton}, the
assignment $V=\tuple{x,e,d,e,x,x}$ has a segmentation $\tuple{x,e},
\tuple{e,x,x}$ with $\ell=2$, which is accepted by the automaton via
the string $\tuple{x,e,x,e,x,x}$ with $\alpha_1 = \alpha_3 = \epsilon$
(the empty string) and $\alpha_2 = \tuple{d}$.

Given an assignment, the algorithm presented below initialises and
updates a segmentation. The violations of the constraint and its
decision variables are calculated relative to the current
segmentation:

\begin{definition}[Violations]
  \label{def:viol}
  Given an automaton describing a constraint $c$ and given a segmentation
  $\sigma_1,\dots,\sigma_\ell$ of an assignment for a sequence of $n$
  decision variables $V_1,\dots,V_n$:
  \begin{itemize}
  \item The \emph{constraint violation} of $c$ is
    $n - \sum_{j=1}^\ell |\sigma_j|$.
  \item The \emph{variable violation} of decision variable $V_i$ is
    $0$ if there exists a segment index $j$ in $1,\dots,\ell$ such that
    $i \in \sigma_j$, and $1$ otherwise.
  \end{itemize}
\end{definition}

It can easily be seen that the violation of a constraint is also the
sum of the violations of its decision variables, and that it is never
an underestimate of the minimal Hamming distance between the current
assignment and any satisfying assignment.

Our approach, described in the next three sub-sections, greedily grows
a segmentation from left to right across the current assignment
relative to a satisfying assignment, and makes stochastic choices
whenever greedy growth is impossible.

\subsection{Initialisation}
\label{sect:init}

A finite automaton is first unrolled for a given length $n$ of a
sequence $V=\tuple{V_1,\dots,V_n}$ of decision variables, as in
\cite{Pesant:seqs}:

\begin{definition}[Layered Graph]
  Given a finite automaton with $m$ states, the \emph{layered graph}
  over a given number $n$ of decision variables is a graph with $m
  \cdot (n+1)$ nodes. Each of the $n+1$ vertical layers has a node for
  each of the $m$ states of the automaton. The node for the start
  state of the automaton in layer~1 is marked as the start node.
  There is an arc labelled~$w$ from node~$f$ in layer~$i$ to node~$t$
  in layer~$i+1$ if and only if there is a transition labelled~$w$
  from~$f$ to~$t$ in the automaton. A node in layer~$n+1$ is marked
  as a success node if it corresponds to a success state in the
  automaton.
\end{definition}

\begin{figure}[t]
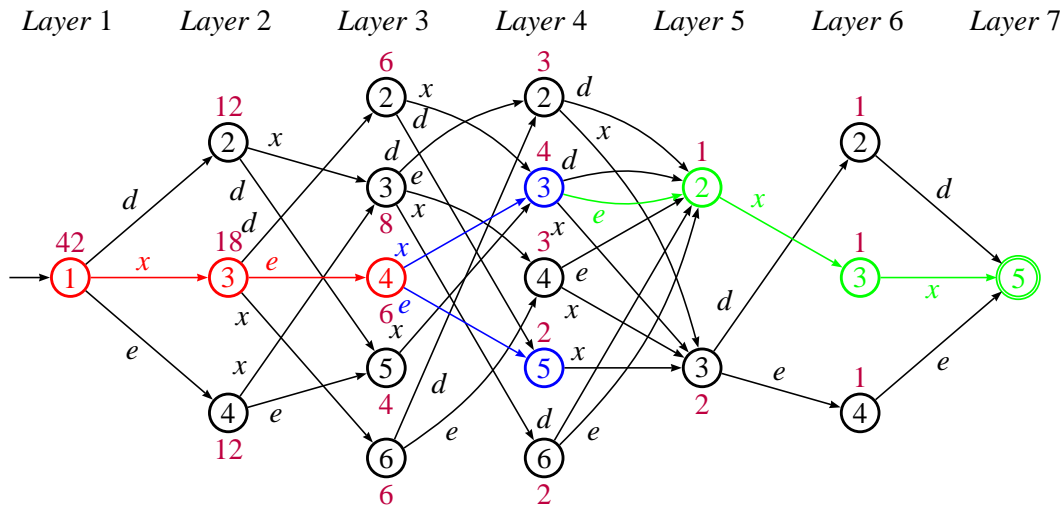

  \centering
  \VCDraw[1]{%
    \begin{VCPicture}{(0,-1)(22,11)}   
      \ChgStateLineColor{white} \ChgStateLabelColor{white}
      \State{(-1,9.2)}{a1} \State{(1,9.2)}{a2}
      \State{(2.5,9.2)}{b1} \State{(4.5,9.2)}{b2}
      \State{(6,9.2)}{c1} \State{(8,9.2)}{c2}
      \State{(9.5,9.2)}{d1} \State{(11.5,9.2)}{d2}
      \State{(13,9.2)}{e1} \State{(15,9.2)}{e2}
      \State{(16.5,9.2)}{f1} \State{(18.5,9.2)}{f2}
      \State{(20,9.2)}{g1} \State{(22,9.2)}{g2}
      \ChgEdgeLineColor{white}
      \EdgeL{a1}{a2}{Layer~1}
      \EdgeL{b1}{b2}{Layer~2}
      \EdgeL{c1}{c2}{Layer~3}
      \EdgeL{d1}{d2}{Layer~4}
      \EdgeL{e1}{e2}{Layer~5}
      \EdgeL{f1}{f2}{Layer~6}
      \EdgeL{g1}{g2}{Layer~7}
      \ChgEdgeLineColor{black}
      
      \ChgStateLineColor{white} \ChgStateLabelColor{purple}
      \State[1]{(17.5,7.8)}{b} \State[1]{(17.5,4.8)}{c} \State[1]{(17.5,1.8)}{d}
      \State[1]{(14,6.8)}{e} \State[2]{(14,1.2)}{f}
      \State[3]{(10.5,8.8)}{g} \State[4]{(10.5,6.8)}{h} \State[3]{(10.5,4.8)}{i}
      \State[2]{(10.5,2.8)}{j} \State[2]{(10.5,-0.8)}{k}
      \State[6]{(7,8.8)}{l} \State[8]{(7,5.2)}{m} \State[6]{(7,3.2)}{n}
      \State[4]{(7,1.2)}{o} \State[6]{(7,-0.8)}{p}
      \State[12]{(3.5,7.8)}{q} \State[18]{(3.5,4.8)}{r} \State[12]{(3.5,0.2)}{s}
      \State[42]{(0,4.8)}{t}
      
      \ChgStateLineColor{black} \ChgStateLabelColor{black}
      \State[2]{(3.5,7)}{22}  \State[4]{(3.5,1)}{24}
      \State[2]{(7,8)}{32}  \State[3]{(7,6)}{33} \State[5]{(7,2)}{35} \State[6]{(7,0)}{36}
      \State[2]{(10.5,8)}{42} \State[4]{(10.5,4)}{44} \State[6]{(10.5,0)}{46}
      \State[3]{(14,2)}{53}      
      \State[2]{(17.5,7)}{62} \State[4]{(17.5,1)}{64}
         
      \ChgStateLineColor{red}\ChgStateLabelColor{red}
      \State[1]{(0,4)}{11} \State[3]{(3.5,4)}{23} \State[4]{(7,4)}{34}

      \ChgStateLineColor{green}\ChgStateLabelColor{green}
      \State[2]{(14,6)}{52} \State[3]{(17.5,4)}{63} \FinalState[5]{(21,4)}{75}
     
      \ChgStateLineColor{blue}\ChgStateLabelColor{blue}
      \State[3]{(10.5,6)}{43} \State[5]{(10.5,2)}{45}
      
      \Initial{11}
            
      \EdgeL{11}{22}{d}  \EdgeR{11}{24}{e}     
      \EdgeR[0.1]{22}{35}{d} \EdgeL[0.2]{22}{33}{x}
      \EdgeR[0.1]{23}{36}{x} \EdgeL[0.17]{23}{32}{d}
      \EdgeR[0.2]{24}{35}{e} \EdgeL[0.1]{24}{33}{x}       
      \EdgeL[0.06]{32}{45}{d} \ArcL[0.1]{32}{43}{x}
      \ArcL[0.04]{33}{44}{e} \EdgeL[0.06]{33}{46}{x} \ArcL[0.05]{33}{42}{d}
      \EdgeL[0.07]{35}{43}{x}
      \EdgeR[0.2]{36}{42}{d} \ArcR[0.21]{36}{44}{e}     
      \ArcL[0.1]{42}{52}{d} \ArcL[0.15]{42}{53}{x}
      \EdgeR[0.1]{43}{53}{x} \ArcL[0.08]{43}{52}{d}
      \EdgeR[0.15]{44}{53}{x} \EdgeR[0.1]{44}{52}{e}
      \EdgeL[0.12]{45}{53}{x}
      \EdgeL[0.05]{46}{52}{d} \ArcR[0.1]{46}{52}{e}          
      \EdgeL{53}{64}{e}  \EdgeL[0.2]{53}{62}{d}     
      \EdgeL{62}{75}{d} 
      \EdgeR{64}{75}{e}
      \ChgEdgeLineColor{red}\ChgEdgeLabelColor{red}
      \EdgeL{11}{23}{x} \EdgeL[0.2]{23}{34}{e} 
      \ChgEdgeLineColor{green}\ChgEdgeLabelColor{green}
      \ArcR[0.30]{43}{52}{e} \EdgeL[0.25]{52}{63}{x} \EdgeR{63}{75}{x}
      \ChgEdgeLineColor{blue}\ChgEdgeLabelColor{blue}
      \EdgeL[0.05]{34}{43}{x} \EdgeR[0.07]{34}{45}{e}     
    \end{VCPicture}
  }  
  \caption{The minimised unrolled automaton of Figure~\ref{fig:automaton}.  The
    number by each node is the number of paths from that node
    to the success node in the last layer. The colour coding is purely
    for the convenience of the reader to spot a particular path
    mentioned in the running text.}
  \label{fig:unroll}
\end{figure}

The layered graph is further processed by removing all nodes and arcs
that do not lead to a success node. The resulting graph, seen as a DFA
(or as an ordered MDD), need \emph{not} be minimised (or reduced) for
our approach (although this is a good idea for the global search
approaches~\cite{Beldiceanu:automata,Pesant:seqs}, as argued
in~\cite{Lagerkvist:licentiate}, and would be a good idea for the
local search approach of~\cite{Pralong:regular}), as the number of
arcs of the graph does not influence the time complexity of our
algorithm below. For instance, the minimised unrolled version
for $n=6$ decision variables of the automaton in
Figure~\ref{fig:automaton} is given in Figure~\ref{fig:unroll}. Note
that a satisfying assignment $\tuple{d_1,\dots,d_n}$ corresponds to a
path from the start node in layer $1$ to a success node in layer
$n+1$, such that each arc from layer $i$ to layer $i+1$ of this path
is labelled $d_i$.

Further, we require a number of data structures, where $m$ is the
number of states in the given automaton and $n$ is the number of
decision variables it was unrolled for:
\begin{itemize}
\item $nbrPaths[1 \leq i \leq n, 1 \leq j \leq m]$ records the
  number of paths from node $j$ in layer $i$ to a success node in the
  last layer; for example, see the numbers by each node in
  Figure~\ref{fig:unroll};
\item $\ell$ is the number of segments in the current segmentation;
\item segments $\sigma_1,\dots,\sigma_{\ell}$ record the current
  segmentation;
\item $\Vio[1 \leq i \leq n]$ records the current violation of decision
  variable $V_i$ (see Definition~\ref{def:viol});
\end{itemize}
The $nbrPaths$ matrix can be computed in straightforward fashion by
dynamic programming. The other three data structures are initialised
(when the starting position is $s=1$) and maintained (when decision
variable $V_s$ is changed, with $s \geq 1$) by the $\calcSegment(s)$
procedure of Algorithm~\ref{algo:calcSegment}. Upon some
initialisations (lines~2 and~3), it (re)visits only the decision
variables $V_s,\dots,V_n$ (line~4). If the value of the currently
visited decision variable $V_i$ triggers the extension of the
currently last segment (lines~6 and~9) or the creation of a new
segment (lines~6 to~9), then its violation is $0$ (line~10).
Otherwise, its violation is $1$ and a successor node is picked with a
probability weighted according to the number of paths from the current
node to a success node (lines~11 to~14). Toward this, we maintain the
nodes of the picked path (line~16).

\begin{algorithm}[t]
  \begin{algorithmic}[1]
    \Procedure $\calcSegment(s: 1,\dots,n)$
    \STATE let $\ell$ be the number of segments picked for
    $\tuple{V_1,\dots,V_{s-1}}$ at the previous run; assume $\ell=0$
    at the first run
    \STATE $node[1] \IsAssigned 1$; $inSegment \IsAssigned \True$
    \FORALL{$i \IsAssigned s$ \To $n$}
    \IF{the current value, say $a$, of $V_i$ is the label of an arc
      from $node[i]$ to some node $t$}
    \IF{$\Not inSegment$}
    \STATE $\ell \IsAssigned \ell+1$;
    $\sigma_\ell \IsAssigned \epsilon$;
    $inSegment \IsAssigned \True$
    \COMMENT create a new segment
    \ENDIF
    \STATE $\sigma_\ell \IsAssigned \sigma_\ell \cdot a$
    \STATE $\Vio[i] \IsAssigned 0$
    \ELSE
    \STATE $inSegment \IsAssigned \False$
    \STATE $\Vio[i] \IsAssigned 1$
    \STATE pick a successor $t$ of $node[i]$ with probability
    $nbrPaths[i+1,t] ~/~ nbrPaths[i,node[i]]$
    \ENDIF
    \STATE $node[i+1] \IsAssigned t$
    \ENDFOR
  \end{algorithmic}
  \caption{Computation and update of the current segmentation from
    position $s$}
  \label{algo:calcSegment}
\end{algorithm}

The time complexity of Algorithm~\ref{algo:calcSegment} is linear in
the number $n$ of decision variables, because only one path (from
layer $s$ to layer $n+1$) is explored, with a constant-time effort at
each node. Once the pre-processing is done, the time complexity of
Algorithm~\ref{algo:calcSegment} is thus \emph{independent} of the
number of arcs of the unrolled automaton! Hence the minimisation (or
reduction) of the unrolled automaton would be merely for space savings
(and for the convenience of human reading) as well as for accelerating
the pre-processing computation of the $nbrPaths$ matrix. In our
experiments, these space and time savings are not warranted by the
time required for minimisation (or reduction).

Note that this algorithm works \emph{without} change or loss of
performance on \emph{non}-deterministic finite automata (NFAs). This
is potentially interesting since NFAs are often smaller than their
equivalent DFAs, but (as just seen) the number of arcs has no
influence on the time complexity of Algorithm~\ref{algo:calcSegment}.

For example, in Figure~\ref{fig:unroll}, with the initial assignment
$V = \tuple{x,e,d,e,x,x}$ and a first call to
Algorithm~\ref{algo:calcSegment} with $s=1$, the first segment will be
$\tuple{x,e}$ (the red path). Next, the assignment $V_3=d$ triggers a
violation of $1$ for decision variable $V_3$ (we say that it is a
\emph{violated variable}) because there is no arc labelled $d$ that
connects the current node~4 in layer~3 with any nodes in layer~4.
However, node~4 in layer~3 has two out-going arcs, namely to nodes~3
and~5 in layer~4 (in blue). In layer~4, there are $4$ paths from
node~3 to the last layer, compared to $2$ such paths from node~5, so
node~3 is picked with probability $\frac{4}{6}$ and node~5 is picked
with probability $\frac{2}{6}$ (where the $2$, $4$, and $6$ are the
purple numbers by those nodes), and we assume that node~3 in layer~4
is picked. From there, we get the second segment~$\tuple{e,x,x}$ (the
green path), which stops at success node~5 in the last layer. The
violation of the constraint is thus $1$, because the value of one
decision variable does not participate in any segment.

Continuing the example, we assume now that decision variable $V_3$ is
changed to value $e$, and hence we call
Algorithm~\ref{algo:calcSegment} with $s=3$. Only $\ell=1$ segment can
be kept from the previous segmentation picked for $\tuple{V_1,V_2}$,
namely $\tuple{x,e}$ (the red path). Since there is an arc labelled
$e$ from the current node~4 in layer~3, namely to node~5 in layer~4,
segment~$\ell$ is extended (line~9) to $\tuple{x,e,e}$. However, with
decision variable $V_4$ still having value $e$, this segment cannot be
extended further, since there is no arc labelled $e$ from node~5 in
layer~4, and hence $V_4$ is violated. Similarly, decision variables
$V_5=x$ and $V_6=x$ are violated no matter which successors are
picked, so no new segment is ever created. The violation of the
constraint is thus $3$ because the value of three decision variables
do not participate in any segment. Hence changing decision variable
$V_3$ from value $d$ to value $e$ would not be considered a good local
move, as the constraint violation increases from~$1$ to~$3$. Changing
decision variable $V_3$ to value $x$ instead would be a much better
local move, as the first segment $\tuple{x,e}$ is then extended to the
entire current assignment $\tuple{x,e,d,e,x,x}$, without detecting any
violated variables, so that the violation of the constraint is then
$0$, meaning that a satisfying assignment was found.

In Section~\ref{sect:exp}, we experiment with a deterministic
method~\cite{Pralong:regular} for picking the next node and
experimentally show that our random pick is computationally quicker at
finding solutions.

\subsection{Differentiability}

At present, the differences of the (constraint and variable)
violations upon a candidate local move are calculated na\"ively by
first making the candidate move and then undoing it.

\subsection{Incrementality}

Local search proceeds from the current assignment by checking a number
of neighbours of that assignment and picking a neighbour that ideally
reduces the violation: the exact heuristics are often problem
dependent. But in order to make local search computationally
efficient, the violations of the constraint and its decision variables
have to be computed in an incremental fashion whenever a decision
variable changes value. As shown in Subsection~\ref{sect:init}, our
initialisation Algorithm~\ref{algo:calcSegment} can also be invoked
with an arbitrary starting position $s$ when decision variable~$V_s$
is assigned a new value.

We have implemented this algorithm in \emph{Comet} \cite{Comet:book},
an object-oriented CP language with among others a CBLS back-end
(available at \url{www.dynadec.com}).

\section{Experiments}
\label{sect:exp}

We now establish the practicality of the proposed violation
maintenance algorithm by experimenting with it.  All local search
experiments were conducted under \emph{Comet} (version~2.0 beta) on
an Intel 2.4 GHz Linux machine with 512 MB memory while the constraint programming examples where implemented using SICStus Prolog.

\begin{table}[t]
  \begin{center}
    \begin{tabular}{c|ccccccc}
      & Mon & Tue & Wed & Thu & Fri & Sat & Sun \\
      \hline
      1&$x$&$x$&$x$&$d$&$d$&$d$&$d$\\
      2&$x$&$x$&$e$&$e$&$e$&$x$&$x$\\
      3&$d$&$d$&$d$&$x$&$x$&$e$&$e$\\
      4&$e$&$e$&$x$&$x$&$n$&$n$&$n$\\
      5&$n$&$n$&$n$&$n$&$x$&$x$&$x$\\
    \end{tabular}
  \end{center}
  \caption{A five-week rotating schedule with uniform daily workload $(1d,1e,1n,2x)$}
  \label{tab:rotSched}
\end{table}

Many industries and services need to function around the clock.
Rotating schedules such as the one in Table~\ref{tab:rotSched} (a
real-life example taken from~\cite{Laporte:rotating}) are a popular
way of guaranteeing a maximum of equity to the involved work teams
(see \cite{Laporte:rotating}). In our first benchmark, there are day
($d$), evening ($e$), and night ($n$) shifts of work, as well as days
off ($x$). Each team works maximum one shift per day. The scheduling
horizon has as many weeks as there are teams. In the first week,
team~$i$ is assigned to the schedule in row~$i$. For any next week,
each team moves down to the next row, while the team on the last row
moves up to the first row. Note how this gives almost full equity to
the teams, except, for instance, that team~$1$ does not enjoy the six
consecutive days off that the other teams have, but rather three
consecutive days off at the beginning of week~$1$ and another three at
the end of week~$5$. The daily workload may be uniform: for instance,
in Table~\ref{tab:rotSched}, each day has exactly one team on-duty for
each work shift, and two teams entirely off-duty; we denote this as
$(1d,1e,1n,2x)$; assuming the work shifts average $8$h, each employee
will work $7 \cdot 3 \cdot 8 = 168$h over the five-week-cycle, or
$33.6$h per week. Daily workload, whether uniform or not, can be
enforced by global cardinality ($\GCC$) constraints~\cite{Regin:gcc}
on the columns. Further, any number of consecutive workdays must be
between two and seven, and any change in work shift can only occur
after two to seven days off. This can be enforced by a
$\Pattern(X,\{(d,x),(e,x),(n,x),(x,d),(x,e),(x,n)\})$
constraint~\cite{Pesant:pattern} and a circular
$\Stretch(X,[d,e,n,x],[2,2,2,2],[7,7,7,7])$
constraint~\cite{Pesant:stretch} on the table flattened row-wise into
a sequence~$X$.

Our model posts the $\Pattern$ and $\Stretch$ constraints described by
automata. The $\GCC$ constraints on the columns of the matrix are kept
invariant: the first assignment is chosen so as to satisfy them, and
then only swap moves inside a column are considered. As a
meta-heuristic, we use tabu search with restarting. At each iteration,
the search procedure in Algorithm~\ref{alg:search} selects a violated
variable $x$ (line~4; recall that the violation of a decision variable
is here at most $1$) and another variable $y$ of distinct value in the
same column so that their swap (line~8) gives the greatest violation
change (lines~5 to~7). The length of the tabu list is the maximum
between $6$ and the sum of the violations of all constraints (lines~9
and~10). The best solution so far is maintained (lines~11 to~14).
Restarting is done every $2 \cdot |X|$ iterations (lines~15 and~16).
The expressions for the length of the tabu list and the restart
criterion were experimentally determined.

\begin{algorithm}[t]
\begin{verbatim}
 1: void search(var{int}[] V, ConstraintSystem<LS> S, var{int} violations,
 2:             Solution bestSolution, Counter it, int best, int[,] tabu,
 3:             int restartIter){
 4:   select(x in 1..n : S.violation(V[x]) > 0)
 5:     selectMin(y in 1..n : (x-y) % 7 == 0 && V[x] != V[y],
 6:               nv = S.getSwapDelta(V[x],V[y]) :
 7:               tabu[x,y] <= it || (violations+nv) < best)(nv){
 8:       V[x] :=: V[y];
 9:       tabu[x,y] = it + max(violations,6);
10:       tabu[y,x] = tabu[x,y];
11:       if(best > violations){
12:         best = violations;
13:         bestSolution = new Solution(ls);
14:       }
15:       it++;
16:       if(it % restartIter == 0) restart();
17:     }
18: }
\end{verbatim}
  \caption{The search procedure}
  \label{alg:search}
\end{algorithm} 

Recall that Algorithm~\ref{algo:calcSegment} computes a greedy random
segmentation; hence it might give different segmentations when used for
probing a swap and when used for actually performing that swap.
Therefore, we record the segmentation of each swap probe, and at the
actual swap we just apply its recorded segmentation.

We ran experiments over the eight instances from $(2d,1e,1n,2x)$ to
$(16d,8e,8n,16x)$ (we write the latter as $(2d,1e,1n,2x) \cdot 8$)
with uniform daily workload, where the weekly workload is $37.3$h. For
example, instance $(2d,1e,1n,2x)\cdot3$ has the uniform daily workload
of $2\cdot3$ teams on the day shift, $1\cdot3$ teams on the evening
shift, $1\cdot3$ teams on the night shift, and $2\cdot3$ teams
off-duty. Table~\ref{tab:exp:autrand} gives statistics on the run
times and numbers of iterations to find the first solutions over $100$
runs from random initial assignments.

\begin{table}[t]
  \begin{center}
    \begin{tabular}{|l|r|r|r|r|r|r|r|r|} 
      \hline
      & \multicolumn{4}{c|}{optimisation time (ms)} & \multicolumn{4}{c|}{number of iterations} \\
      \cline{2-9}
      instance & min  & max & avg & $\sigma$ & min  & max & avg & $\sigma$ \\
      \hline
      $(2d,1e,1n,2x)\cdot1$ &  6 & 100 & 22 & 20 & 9 & 528 & 115 & 108 \\
      $(2d,1e,1n,2x)\cdot2$ &  12 &  692 &  168 & 154 & 32 & 2484 & 585 & 561 \\
      $(2d,1e,1n,2x)\cdot3$ &  32 & 2588 & 688 & 611 & 44 & 6612 & 1726 & 1571 \\
      $(2d,1e,1n,2x)\cdot4$ & 80 & 6553 & 1199 & 1275 & 86 & 11212 & 2125 & 2303 \\
      $(2d,1e,1n,2x)\cdot5$ & 60 & 9373 & 1417 & 1545 & 72 & 15604 & 2292 & 2556 \\
      $(2d,1e,1n,2x)\cdot6$ & 160 & 5901 & 1527 & 1227 & 161 & 8051 & 2051 & 1681 \\
      $(2d,1e,1n,2x)\cdot7$ & 176 & 9896 & 1720 & 1686 & 157 & 11680 & 1966 & 1981 \\
      $(2d,1e,1n,2x)\cdot8$ & 216 & 12472 & 2620 & 2309 & 150 & 12588 & 2603 & 2354 \\
      \hline
    \end{tabular}
  \end{center}
  \caption{Minimum, maximum, average, standard deviation of optimisation times (in milliseconds) and numbers of iterations to the first solutions of rotating nurse schedules (100~runs) from \emph{random} initial assignments.}
  \label{tab:exp:autrand}
\end{table}

Posting the product of the $\Pattern$ and $\Stretch$ automata
(accepting the intersection of their two regular languages) has been
experimentally determined to be more efficient than posting the two
automata individually, hence all experiments in this paper use the
product automaton.

Further improvements can be achieved by using a non-random initial
assignment. Table~\ref{tab:exp:autnonrand} gives statistics on the
run times and numbers of iterations to find the first solutions over
$100$ runs, where the initial assignment of instance
$(2d,1e,1n,2x)\cdot i$ consists of $i$ copies of
Table~\ref{tab:nonrandom_assignment}. The results show that this
non-random initialisation provides a better starting point. Although
much more experimentation is required, these initial results show that
even on the instance $(2d,1e,1n,2x)\cdot8$ with $336$ decision
variables it is possible to find solutions quickly.

\begin{table}[t]
  \begin{center}
    \begin{tabular}{|l|r|r|r|r|r|r|r|r|} 
      \hline
      & \multicolumn{4}{c|}{optimisation time (ms)} & \multicolumn{4}{c|}{number of iterations} \\
      \cline{2-9}
      instance & min  & max & avg & $\sigma$ & min & max & avg & $\sigma$ \\
      \hline
      $(2d,1e,1n,2x)\cdot1$ &  1 & 16 & 2 & 3 & 6 & 61 & 11 & 9 \\
      $(2d,1e,1n,2x)\cdot2$ &  1 &  64 &  12 & 12 & 13 & 235 & 34 & 46 \\
      $(2d,1e,1n,2x)\cdot3$ &  12 & 76 & 25 & 13 & 21 & 173 & 46 & 29 \\
      $(2d,1e,1n,2x)\cdot4$ & 28 & 172 & 51 & 27 & 32 & 297 & 72 & 49 \\
      $(2d,1e,1n,2x)\cdot5$ & 28 & 200 & 79 & 42 & 34 & 286 & 106 & 61 \\
      $(2d,1e,1n,2x)\cdot6$ & 56 & 368 & 135 & 76 & 61 & 487 & 156 & 101 \\
      $(2d,1e,1n,2x)\cdot7$ & 84 & 768 & 188 & 123 & 69 & 848 & 189 & 140 \\
      $(2d,1e,1n,2x)\cdot8$ & 112 & 764 & 233 & 112 & 72 & 736 & 202 & 113 \\
      \hline
    \end{tabular}
  \end{center}
  \caption{Minimum, maximum, average, standard deviation of optimisation times (in milliseconds) and numbers of iterations to the first solutions of rotating nurse schedules (100~runs) from \emph{non-random} initial assignments.}
  \label{tab:exp:autnonrand}
\end{table}

\begin{table}[t]
  \begin{center}
    \begin{tabular}{c|ccccccc}
      & Mon & Tue & Wed & Thu & Fri & Sat & Sun \\
      \hline
      1&$d$&$d$&$d$&$d$&$d$&$d$&$d$\\
      2&$e$&$e$&$e$&$e$&$e$&$e$&$e$\\
      3&$n$&$n$&$n$&$n$&$n$&$n$&$n$\\
      4&$x$&$x$&$x$&$x$&$x$&$x$&$x$\\
      5&$d$&$d$&$d$&$d$&$d$&$d$&$d$\\
      6&$x$&$x$&$x$&$x$&$x$&$x$&$x$\\
    \end{tabular}
  \end{center}
  \caption{Non-random initial assignment for the instance $(2d,1e,1n,2x)$}
  \label{tab:nonrandom_assignment}
\end{table}

The only related work we are aware of is a \emph{Comet} implementation
\cite{Pralong:regular} of the $\Regular$ constraint
\cite{Pesant:seqs}, based on the ideas for the propagator of the soft
$\Regular$ constraint \cite{vanHoeve:soft}. The difference is that
they estimate the violation change compared to the \emph{nearest}
solution (in terms of Hamming distance from the current assignment),
whereas we estimate it compared to \emph{one} randomly picked
solution. In our terminology (although it is not implemented that way
in \cite{Pralong:regular}), they find a segmentation, such that an
accepting string for the automaton has the minimal Hamming distance to
the current assignment.

Tables~\ref{tab:exp:aut_can_cp_ran} and~\ref{tab:exp:aut_can_cp_unran}
give comparisons between (our re-implementation of)
$\Regular$~\cite{Pralong:regular}, our method, and a SICStus Prolog
constraint program (CP) where the product automaton of the $\Pattern$
and $\Stretch$ constraints was posted using the built-in
propagation-based implementation of the $\Automaton$ constraint
\cite{Beldiceanu:automata}. These experiments show:
\begin{itemize}
\item Compared with $\Regular$~\cite{Pralong:regular}, our method has
  a higher number of iterations, as it is more stochastic. However,
  our run times are lower, as our cost of one iteration is much
  smaller (linear in the number of decision variables, instead of
  linear in the number of arcs of the unrolled automaton).
\item Compared with the CP method, both local search methods need more
  time to find the first solution when the number of weeks is small.
  However, when the number of weeks increases, the runtime of CP
  increases sharply. From the instance $(2d,1e,1n,2x)\cdot7$, the
  runtime of CP exceeds the average runtime of our method. From the
  instance $(2d,1e,1n,2x)\cdot8$, the runtime of CP exceeds also the
  average runtime of $\Regular$~\cite{Pralong:regular}.
\end{itemize}

\begin{table}[t]
  \begin{center}
    \begin{tabular}{|l|r|r|r|r|r|r|r|r|r|r|} 
      \hline
       & \multicolumn{5}{c|}{optimisation time (ms)} & \multicolumn{4}{c|}{number of iterations} \\
      \cline{2-10}
      instance & \multicolumn{2}{c|}{our method} & \multicolumn{2}{c|}{$\Regular$~\cite{Pralong:regular}} & \multicolumn{1}{c|}{CP} & \multicolumn{2}{c|}{our method} & \multicolumn{2}{c|}{$\Regular$~\cite{Pralong:regular}} \\
      \cline{2-5}  \cline{7-10}
       & avg  & $\sigma$ & avg  & $\sigma$ & & avg  & $\sigma$ & avg  & $\sigma$ \\
      \hline
      $(2d,1e,1n,2x)\cdot1$ &  22 & 20 & 395 & 378 & 10 & 115 & 108 & 98 & 100 \\
      $(2d,1e,1n,2x)\cdot2$ &  168 & 154 & 1584 & 1187 & 10 & 585 & 561 & 223 & 170\\
      $(2d,1e,1n,2x)\cdot3$ &  688 & 611 & 3441 & 2871 & 10 & 1726 & 1571& 333 & 287\\
      $(2d,1e,1n,2x)\cdot4$ & 1199 & 1275 & 5584 & 4423 & 40 & 2125 & 2303 & 399 & 319 \\
      $(2d,1e,1n,2x)\cdot5$ & 1417 & 1545 & 8828 & 7606 & 100 & 2292 & 2556 & 514 & 444 \\
      $(2d,1e,1n,2x)\cdot6$ & 1527 & 1227 & 13888 & 10863 & 510 & 2051 & 1681 & 672 & 529  \\
      $(2d,1e,1n,2x)\cdot7$ & 1720 & 1686 & 13170 & 9814 & 3520 & 1966 & 1981 & 536 & 485 \\
      $(2d,1e,1n,2x)\cdot8$ & 2620 & 2309 & 20202 & 11530 & 25820 & 2603 & 2354 & 745 & 602 \\
      St~Louis~Police & 12740 & 11199 & 50261 & 48026 & -- & 20287 & 17952 & 3248 & 2498  \\
      \hline
    \end{tabular}
  \end{center}
  \caption{Comparison between our method, $\Regular$~\cite{Pralong:regular}, and a SICStus Prolog program using the $\Automaton$~\cite{Beldiceanu:automata} constraint: average and standard deviation of optimisation times (in milliseconds) and numbers of iterations to the first solutions; rotating nurse schedules (100~runs) and the St Louis Police instance (50~runs), from \emph{random} initial assignments.}
  \label{tab:exp:aut_can_cp_ran}
\end{table}

\begin{table}[t]
  \begin{center}
    \begin{tabular}{|l|r|r|r|r|r|r|r|r|r|} 
      \hline
      & \multicolumn{4}{c|}{optimisation time (ms)} & \multicolumn{4}{c|}{number of iterations} \\
      \cline{2-9}
      instance & \multicolumn{2}{c|}{our method} & \multicolumn{2}{c|}{$\Regular$~\cite{Pralong:regular}} & \multicolumn{2}{c|}{our method} & \multicolumn{2}{c|}{$\Regular$~\cite{Pralong:regular}} \\
      \cline{2-9}
       & avg  & $\sigma$ & avg  & $\sigma$ & avg  & $\sigma$ & avg  & $\sigma$ \\
      \hline
      $(2d,1e,1n,2x)\cdot1$ &  2 & 3 & 44 & 26 & 11 & 9 & 10 & 7 \\
      $(2d,1e,1n,2x)\cdot2$ &  12 & 12 & 182 & 65 & 34 & 46 & 22 & 9\\
      $(2d,1e,1n,2x)\cdot3$ &  25 & 13 & 478 & 201 & 46 & 29& 41 & 17\\
      $(2d,1e,1n,2x)\cdot4$ & 51 & 27 & 834 & 254 & 72 & 49 & 56 & 18 \\
      $(2d,1e,1n,2x)\cdot5$ & 79 & 42 & 1546 & 876 & 106 & 61 & 87 & 51 \\
      $(2d,1e,1n,2x)\cdot6$ & 135 & 76 & 2414 & 1233 & 156 & 101 & 113 & 59  \\
      $(2d,1e,1n,2x)\cdot7$ & 188 & 123 & 4517 & 3276 & 189 & 140 & 181 & 134 \\
      $(2d,1e,1n,2x)\cdot8$ & 233 & 112 & 4473 & 1958 & 202 & 113 & 160 & 71 \\
      St~Louis~Police & 3990 & 4012 & 67159 & 55632 & 5389 & 5598 & 3949 & 3159  \\
      \hline
    \end{tabular}
  \end{center}
  \caption{Comparison between our method and $\Regular$~\cite{Pralong:regular}: average and standard deviation of optimisation times (in milliseconds) and numbers of iterations to the first solutions; rotating nurse schedules (100~runs) and the St Louis Police instance (50~runs), from \emph{non-random} initial assignments.}
  \label{tab:exp:aut_can_cp_unran}
\end{table}

Besides the rotating nurse instances, we ran experiments on another,
harder real-life scheduling instance. The St Louis Police problem
(described in \cite{Pralong:regular}) has a seventeen-week-cycle;
however it has more constraints than the rotating nurse problem. It
has non-uniform daily workloads. For example, on Mondays, five teams
work during the day, five at night, four in the evening, and three
teams enjoy a day off; while on Sundays, three teams work during the
day, four at night, four in the evening, and six teams enjoy a day
off. Any number of consecutive workdays must be between three and
eight, and any change in work shift can only occur after two to seven
days off. The problem has other vertical constraints; for example, no
team can work in the same shift on four consecutive Mondays. Further,
the problem has complex $\Pattern$ constraints that limit possible
changes of work shifts; for example, only the patterns $(d,x,d)$,
$(e,x,e)$, $(n,x,n)$, $(d,x,e)$, $(e,x,n)$, and $(n,x,d)$ are allowed.
For this hard real-life problem, our method still works well:
experimental results can also be found in
Tables~\ref{tab:exp:aut_can_cp_ran}
and~\ref{tab:exp:aut_can_cp_unran}.

It is possible to post (see Figure~\ref{fig:cometmodelunrolled}) the
constraints of the rotating nurse problem using the differentiable
invariants \cite{CometInvariants} of \emph{Comet}. This is possible in
general for any automaton by encoding all the paths to a success state
by using \emph{Comet}'s conjunction and disjunction combinators. As
the automata get larger, these expressions can become too large to
post, and even when it is possible to post these expressions our
current experiments show that our approach is more efficient.

\begin{figure}[t]
  \begin{center}  
  \fbox{
    $\begin{array}{l}
      V_1=d \wedge
      \left(
        \begin{array}{l}
          V_2=x \wedge 
          \left(
            \begin{array}{l}
              V_3=d \wedge 
              \left(
                \begin{array}{l}
                  V_4=d \wedge V_5=x \wedge V_6=x \\
                  \vee \\
                  V_4=x \wedge (V_5=d \wedge V_6=d \vee V_5=e \wedge V_6=e)
                \end{array}
              \right) \\  
              \vee \\
              V_3=e \wedge  \dots \\
              \vee \\
              V_3=x \wedge \dots \\            
            \end{array}
          \right) \\             
          \vee \\
          V_2=d \wedge V_3=x \wedge  \dots 
        \end{array}
      \right) \\
      \vee \\
      V_1=x \wedge 
      \left(
        \begin{array}{l}
          V_2=d \wedge \dots \\
          \vee \\
          V_2=e \wedge \dots \\
          \vee \\
          V_2=x \wedge \dots
        \end{array}
      \right) \\
      \vee \\
      V_1=e \wedge 
      \left(
        \begin{array}{l}
          V_2=x \wedge \dots \\
          \vee \\
          V_2=e \wedge \dots 
        \end{array}
      \right)
    \end{array}$
    }
  \end{center}
  \caption{A model for the unrolled DFA in Figure~\ref{fig:unroll} with \emph{Comet} combinators}
  \label{fig:cometmodelunrolled}
\end{figure}

\section{Conclusion}
\label{sect:concl}

In summary, we have shown that the idea of describing novel
constraints by automata can be successfully imported from classical
(global search) constraint programming to constraint-based local
search (CBLS). Our violation algorithms take time linear in the number
of decision variables, whereas the propagation algorithms take
amortised time linear in the number of arcs of the unrolled automaton
\cite{Beldiceanu:automata,Pesant:seqs}. We have also experimentally
shown that our approach is competitive with the CBLS approach
of~\cite{Pralong:regular}.

There is of course a trade-off between using an automaton to describe
a constraint and using a hand-crafted implementation of that
constraint. On the one hand, a hand-crafted implementation of a
constraint is normally more efficient during search, because
properties of the constraint can be exploited, but it may take a lot
of time to implement and verify it. On the other hand, the (violation
or propagation) algorithm processing the automaton is implemented and
verified once and for all, and our assumption is that it takes a lot
less time to describe and verify a new constraint by an automaton than
to implement and verify its algorithm. We see thus opportunities for
rapid prototyping with constraints described by automata: once a
sufficiently efficient model, heuristic, and meta-heuristic have been
experimentally determined with its help, some extra efficiency may be
achieved, if necessary, by hand-crafting implementations of any
constraints described by automata.

As witnessed in our experiments, constraint composition (by conjunction) is
easy to experiment with under the DFA approach, as there exist
standard and efficient algorithms for composing and minimising DFAs,
but there is no known systematic way of composing violation (or
propagation) algorithms when decomposition is believed to obstruct
efficiency.

In the global search approach to CP, the common modelling device of
reification can be used to shrink the size of DFAs describing
constraints \cite{Beldiceanu:automata}. For instance, consider the
$\Element([x_1,\dots,x_n],i,v)$ constraint, which holds if
and only if $x_i=v$. Upon reifying the decision variables
$x_1,\dots,x_n$ into new Boolean decision variables $b_1,\dots,b_n$
such that $x_i=v \Leftrightarrow b_i=1$, it suffices to pose the
$\Automaton([b_1,\dots,b_n],\DFA)$ constraint, where $\DFA$
corresponds to the regular expression $0^*1(0+1)^*$, meaning that at
least one $1$ must be found in the sequence of the $b_i$ decision
variables. However, such explicit reification constraints are not
necessary in constraint-based local search, as a total assignment of
values to all decision variables is maintained at all times: instead
of processing the \emph{values} of the decision variables when
computing the segments, one can process their \emph{reified values}.

It has been shown that the use of counters (initialised at the start
state and evolving during possibly conditional transitions) to enrich
the language of DFAs and thereby shrink the size of DFAs
can be handled in the global search approach to CP
\cite{Beldiceanu:automata}, possibly upon some concessions at the
level of local consistency that can be achieved. In the \emph{Global
  Constraint Catalogue} \cite{GC-catalogue}, some $31$ of the
currently $108$ constraints described by DFAs use lists of counters,
and another $25$ constraints use arrays of counters. We need to
investigate the effects on our violation maintenance algorithm of
introducing counters and conditional transitions.

\subsubsection*{Acknowledgements}

The authors are supported by grant 2007-6445 of the Swedish Research
Council (VR), and Jun He is also supported by grant 2008-611010 of
China Scholarship Council and the National University of Defence
Technology of China. Many thanks to Magnus {\AA}gren (SICS) for some
useful discussions on this work, and to the anonymous referees of both
LSCS'09 and the Doctoral Programme of CP'09, especially for pointing
out the existence of \cite{Pralong:regular,vanHoeve:soft}.

\bibliographystyle{eptcs}

\end{document}